\documentclass{svjour3} [12pt]
 \usepackage{nopageno}

\usepackage{amsfonts}
\usepackage{amsmath}
\usepackage{graphicx}
\usepackage{bm}
\usepackage{natbib}
\setcitestyle{numbers}
\usepackage{xcolor}
\usepackage{caption} 
\usepackage{subcaption}
\usepackage{tikz,forest}
\usetikzlibrary{arrows.meta}
\usepackage{soul}
\usepackage{hyperref}

\setcitestyle{square}

\begin{document}
\pagestyle{plain}

\begin{center}{\Large Leo Breiman, the Rashomon Effect, and the Occam Dilemma}\\
{\large Cynthia Rudin}\\
{\large Duke University}\\
\end{center}


\noindent \textbf{Abstract.} In the famous ``Two Cultures'' paper, Leo Breiman provided a visionary perspective on the cultures of ``data models'' (modeling with consideration of data generation) versus ``algorithmic models'' (vanilla machine learning models). I provide a modern perspective on these two approaches.
One of Breiman's key arguments \textit{against} data models is what he called the ``Rashomon Effect,'' which is the existence of many different-but-equally-good models. The Rashomon Effect implies that data modelers would not be able to determine which model generated the data. Conversely, one of his core advantages \textit{in favor of} data models is simplicity, as he claimed there exists an ``Occam Dilemma,'' i.e., an accuracy-simplicity tradeoff, where algorithmic models must be complex in order to be accurate. After 25 years of more powerful computers, it has become clear that this claim is not generally true, in that algorithmic models do not need to be complex to be accurate; however, there are nuances that help explain Breiman's logic, specifically, that by ``simple,'' he appears to consider only linear models or unoptimized decision trees. Interestingly, the Rashomon Effect is a key tool in proving the nullification of the Occam Dilemma. To his credit though, Breiman did not have the benefit of modern computers, with which my observations are much easier to make.

Breiman's goal for interpretability was somewhat intertwined with causality: simpler models can help reveal which variables have a causal relationship with the outcome. However, I argue that causality can be investigated without the use of single models, whether or not they are simple. Interpretability is useful in its own right, and I think Breiman knew that too. 

Technically, my modern perspective does not belong to either of Breiman's Two Cultures, but shares the goals of both of them -- causality, simplicity, accuracy -- and shows that these goals can be accomplished in other ways, without the limitations Breiman was concerned about.


\section{The Two Cultures: Data Models and Algorithmic Models}

Did Breiman's famous 2001 ``Two Cultures'' paper \cite{breiman2001statistical}  miss the forest for the trees? That question is a play on words -- Breiman made groundbreaking contributions with both classification and regression trees (CART) and random forest. ``Two Cultures'' was brilliant: it revolutionized how the statistical world could understand and gain from the ``algorithmic'' (i.e., machine learning) perspective. 
Specifically, Breiman contrasted the older ``statistical'' way of modeling -- creating a ``data model'' for the data-generation process -- with the newer ``algorithmic'' approach -- creating models that fit the data (e.g., a random forest). The algorithmic models are not designed to represent the data generation process, and can be quite complicated, whereas the data models are often overly simplistic, like plain logistic regression. Data models didn't actually need to fit the data particularly well as long as they revealed which variables were important for data generation.
In his rejoinder, Breiman writes ``I like my division because it is pretty clear cut — are you modeling the inside of the box or not?'' 

Breiman made several arguments for the algorithmic approach, but also discussed what he perceived were its disadvantages.  Breiman's main argument \textit{against} algorithmic models -- that they must be complex and not understandable, which he calls the \textit{Occam Dilemma} -- is not generally true, as we have found after 25 years of machine learning with modern computers. A proof for my claim lies within Breiman's arguments \textit{in favor of} algorithmic models, which is the \textit{Rashomon Effect}, i.e., the existence of many equally good models for the data. Essentially, the Rashomon Effect nullifies the Occam Dilemma.

Specifically, Breiman points out that the Rashomon Effect implies that data modelers could easily come to different conclusions, which means they do not necessarily know anything about the data generation process at all. Algorithmic modelers do not claim to capture data generation, and thus have no such flaw. 

However, Breiman also claims that algorithmic models have a different flaw in that they must be complicated. He writes ``Usually, simple parametric models imposed on data generated by complex systems, for example, medical data, financial data, result in a loss of accuracy and information as compared to algorithmic models.'' Similarly, ``Accuracy generally requires more complex prediction methods. Simple and interpretable functions do not make the most accurate predictors.''  

As I mentioned, the problem with Breiman's statements on simplicity is that they are generally not true.
There are two nuances though. First, Breiman's notion of simplicity is probably different from mine, explaining why he might see a tradeoff whereas I do not.
Second, Breiman misses something important: that the Rashomon Effect does not simultaneously exist with the accuracy/simplicity tradeoff.
In fact, the Rashomon Effect causes the accuracy/simplicity tradeoff (the Occam Dilemma) to vanish. I will discuss this more shortly, but the argument boils down to the basic idea that \textit{if there are many different-but-good models, some of them are likely to be interpretable}.


Generalization, simplicity, causality and computational complexity are key players in Breiman's discussion. Even though many things have changed, the key goals of understanding causality, keeping models interpretable and accurate have endured -- however, I believe there are other ways to achieve these goals than what Breiman considered. For understanding the data generation process, I suggest using model-free methods for variable importance and causal inference rather than Breiman's linear data models. For prediction, I recommend interpretable machine learning methods rather than Breiman's black box algorithmic models.


This perspective accompanies a talk about the paper ``Amazing Things Come from Having Many Good Models'' \cite{RudinEtAlAmazing2024}. 


\section{The Rashomon Effect Nullifies the Occam Dilemma}

Breiman calls ``the conflict between simplicity and accuracy'' the Occam Dilemma. Typically, this is called the accuracy/simplicity or accuracy/interpretability tradeoff.
It has long been observed that there is no accuracy/simplicity tradeoff for a vast set of problems. \cite{Rudin19, RudinEtAlSurvey2022} provide more detail and citations on this topic. For most tabular data problems with noise in the outcomes, very sparse models perform similarly to complicated black box models. 

For several years, my collaborators and I have tried to understand the mathematics of why there is no such tradeoff. 
I will summarize Breiman's definition of the Rashomon Effect, then discuss why the Rashomon Effect does not simultaneously exist with the accuracy/simplicity tradeoff. (Oddly, our ``dilemma'' is to explain why there is no Occam Dilemma.) 

\subsection{The Rashomon Effect}
Breiman coined the Rashomon Effect in the Two Cultures paper, which is the phenomenon that many datasets admit many approximately-equally-good models. He defines it this way: ``What I call the Rashomon Effect is that there is often a multitude of different descriptions [equations f(x)] in a class of functions giving about the same minimum error rate.''
Breiman points out that he is not the first to notice this phenomenon. He quotes McCullagh and Nelder (1989) \cite{McCullaghNelder1989}, who say ``Data will often point with almost equal emphasis on several possible models, and it is important that the statistician recognize and accept this.'' 

The Rashomon Effect is very real, and in recent years, we have been able to enumerate or represent Rashomon sets for various function classes, including decision trees \cite{xin2022exploring} and generalized linear models \cite{zhong2023exploring}. It is very clear that for a wide variety of datasets, there are many very different models that all perform similarly.

As discussed, Breiman uses the Rashomon Effect as an argument against data models: ``Suppose two statisticians, each one with a different approach to data modeling, fit a model to the same data set. Assume also that each one applies standard goodness-of-fit tests, looks at residuals, etc., and is convinced that their model fits the data. Yet the two models give different pictures of nature's mechanism and lead to different conclusions.'' 

\subsection{The Rashomon Set Theory}

I will follow the ``Rashomon Set Theory'' argument of \citep{SemenovaRuPa2022} to show that a large Rashomon Effect (the existence of many good models) gives rise to simpler models that are also accurate. That is, the Rashomon Effect itself suppresses the accuracy/simplicity tradeoff.
Following \citep{SemenovaRuPa2022}, if the Rashomon Effect is large, there are many good models, perhaps including at least one ball of models in some function space. For instance, this could be a ball of polynomials of degree $p$, using an $L_2$ distance metric to describe distances between functions. Intuitively, a ball of functions around a polynomial of degree $p$ includes functions that are smoother than it (unless the function is already linear). If the ball is large enough, the simplest models in the ball may be polynomials of a smaller degree than $p$ -- simpler, more interpretable, functions. 

\begin{figure}[ht]
    \centering
    \includegraphics[width = 0.4\columnwidth]{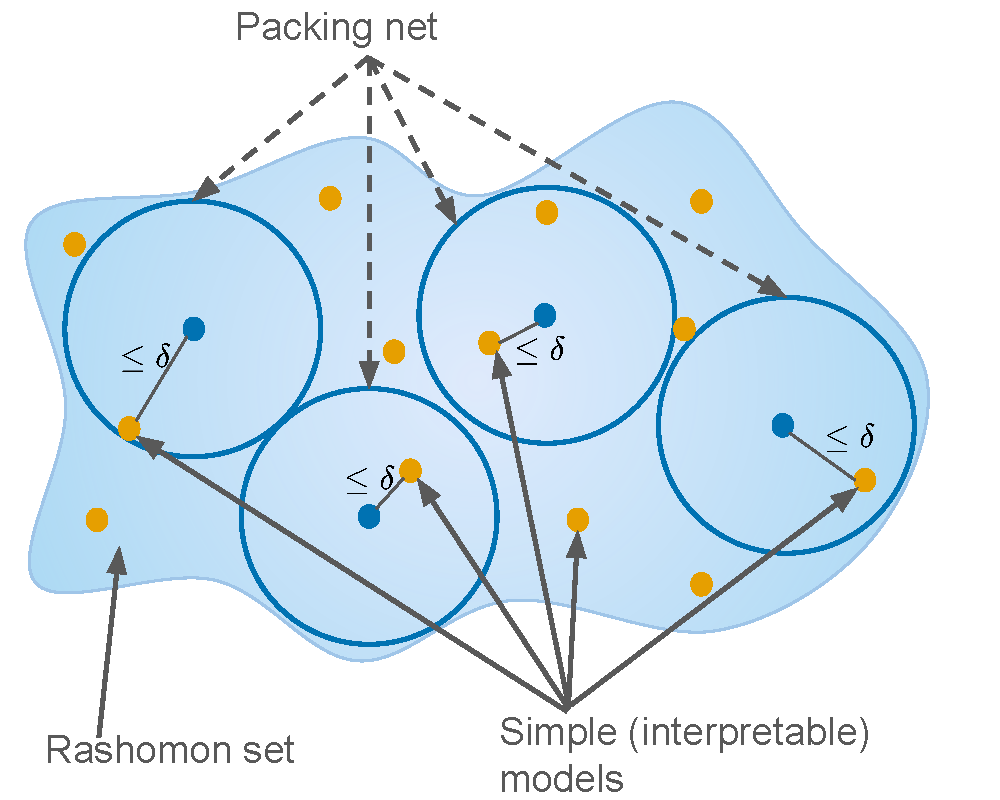}
    \caption{Illustration reproduced from \cite{SemenovaRuPa2022,RudinEtAlAmazing2024}. This shows a Rashomon set (blue region) that is large enough to contain several balls of functions. Function classes are often nested with good approximating properties, so that within each ball of complex models, there is at least one model from a simpler function class within it. So, for every good model in the more complex space (blue region), there exists a $\delta$-close model from the simpler space (orange dots). In this illustration, the Rashomon set contains at least four simpler models, which is its $2\delta$-packing number, where blue dots correspond to the centers of the balls in the packing.}
    \label{fig:rashomon_set_simplicity}
\end{figure}

Thus, we generally would not simultaneously have a large Rashomon Effect and an accuracy/simplicity tradeoff; they are not compatible. Figure \ref{fig:rashomon_set_simplicity} shows an illustration of a large Rashomon Effect giving rise to simpler-but-accurate models.

The Rashomon Set Theory explains that noise in the world causes the Rashomon Effect. We explained this process in \cite{SemenovaEtAl2023, BonerEtAl2024, RudinEtAlAmazing2024} as follows. Noise in the world leads to increased variance in the loss. This increased variance leads to poorer generalization, so the user observes worse test performance. This forces the user to decrease the complexity of the space (which on its own leads to simpler functions), and a larger fraction of models that appear to be about equally good. This, combined with the observation that simple function classes can approximate more complex ones, as illustrated in Figure \ref{fig:rashomon_set_simplicity}, leads to even simpler functions that perform well. 

Based on this logic, as long as there is sufficient noise in the world to produce a Rashomon Effect, and that functions from a simpler class can represent those from the more complex class sufficiently well, the Rashomon Effect does not simultaneously exist with the accuracy/simplicity tradeoff. 

It is important to remark that the Rashomon Effect may not exist for non-noisy problems to the same extent. However, the problems Breiman mentions, like predicting ozone levels and the causes of freeway traffic breakdowns, are noisy -- one would not expect close to perfect prediction accuracy on those problems. He tells us this directly when he writes: ``In two-class data, separability by a hyperplane
does not often occur.'' Let me discuss noisy vs$.$ non-noisy problems in more depth next.

\subsection{The Machine Learning Community Did Not Understand The Occam Dilemma}

Neither the machine learning community nor the statistics community really understood the Occam Dilemma or when it occurs until recently. In practice, ``non-noisy'' problems are completely different from ``noisy'' problems. Non-noisy problems have deterministic labels: given $x$, $y$ is not random. Noisy problems have random $y$ given $x$. Many of Breiman's problems were noisy data problems. But many machine learning researchers worked extensively on non-noisy problems. Computer vision problems, which were being studied extensively by many machine learning researchers, are not noisy -- if an image $x$ contains a chair (the label $y$ is ``chair''), it always contains a chair. I would venture to say that these two problem classes define two unnamed subfields of machine learning, because everything about how we approach the two classes of problems is different. Currently, unimaginably complex neural network models are dominating computer vision and natural language processing, which are non-noisy problems. They haven't made a dent on a multitude of noisy problems -- because they can't. The Rashomon Set Theory explains why not.

Researchers like me, who were working extensively on noisy problems, ended up feeling like the field of machine learning sold us a raw deal: none of its fancy methods performed any better than any other fancy method. In fact, they all performed similarly, thanks to the Rashomon Effect. At the time, academic papers were filled with extensive experimental tables showing tiny little improvements to accuracy scores that researchers had eked out to prove the worth of their new algorithmic idea. Everyone knew that the results were a bit suspect. They were due to some level of experimental manipulation, perhaps using slightly different preprocessing for some of the methods, or running the experiments multiple times to get the desired result. Even if the results were legit, the improvements were so tiny that it wasn't worth implementing the method. (At that time, code wasn't made public for reproducibility.) This was -- honestly -- heartbreaking for a researcher like me, who really wanted to believe that an algorithm could change the world. What I did not know (and others did not know either) is that sophisticated algorithms like neural networks and boosting \textit{could} make a huge impact in non-noisy data problems like computer vision and language, and that a \textit{separate} set of algorithms \textit{could} make a huge impact on noisy problems, namely \textit{interpretable machine learning methods}. For noisy problems, while we could not aim to improve accuracy over other methods thanks to the Rashomon Effect, we could aim to find models that are \textit{massively} more practical in terms of interpretability metrics. These interpretable models could be used in really high stakes decisions whereas black box models could not \cite{Rudin19,RudinRa19}.

I learned this lesson only while working on applications. Starting in 2007, I had been working in power grid maintenance in New York City \cite{RudinEtAl12, RudinEtAl14}. My team's assignment was to predict manhole fires and explosions in NYC. This data was extremely noisy, containing information from trouble tickets typed by dispatchers and underground electrical information dating from the 1880's. No matter what we did, complex methods like boosting and support vector machines produced no improvement over logistic regression. However, producing logistic regression models allowed us to communicate with the power engineers, whose comments helped us troubleshoot the dataset and refine the problem. Interpretability had led to \textit{more} overall accuracy. This allowed me to understand its immense practical value, even if others in the machine learning community did not.

While linear models were about the best one could do for the extremely noisy power reliability problems I worked on, for many other noisy problems, more complexity is needed, but not much more. Sparse GAMs \cite{LiuEtAlFasterRisk2022} and sparse decision trees \cite{lin2020generalized,mctavish2022fast} suffice for most problems. Even simple risk scorecards can perform as well as black box models for many applications if they are optimized properly  \cite{JMLR:v20:18-615,zhu2023GroupFasterRisk,ZengUsRu2017,WangHanEtAl2022, Souillard15}.
However, I believe that Breiman was considering the notion of simplicity narrowly as just linear models, which explains how he could observe what he did, despite the mathematical argument seemingly to the contrary.


\section{There's Simplicity... And Simplicity}

I do not believe Breiman was using the same notion of simplicity as I am when reporting the existence of an accuracy/simplicity tradeoff. My notion of simplicity includes models that are small enough to fit on an index card or piece of paper, like generalized additive models (GAMs) and optimized decision trees.
His notion of simplicity  plausibly included only (linear) logistic regression models and CART trees, which are overly simplistic. In other words, the models I consider simple can be much more powerful than his simple models -- there is a world of difference in predictive power between a linear model and a GAM. For instance, in a linear model that includes the variable ``age,'' the predicted score would need to increase linearly with age. I do not know any criminal justice model or healthcare model where a linear dependence on age is appropriate. For criminal recidivism, we would want the risk to decrease and plateau with age -- this can be done with a GAM. 
Also, as described in Section \ref{subsec:cart}, also a world of difference between a CART tree and an optimized decision tree. 

However, as I will now explain, Breiman may have had little choice but to see simplicity his way, due to both cultural and computational reasons. 

\subsection{Linear Models are Too Simple}

Much of Breiman's exposition is centered on predictive accuracy and generalization. 
Breiman reported seeing simple logistic regression models (models that are too simple to possibly explain the true data generation process) being used to model real-world phenomena, and p-values of their coefficients being used to assess which variables have causal relevance. Breiman writes ``This theory was used
both by academic statisticians and others to derive significance levels for coefficients on the basis of model (R), with little consideration as to whether the data on hand could have been generated by a linear model. Hundreds, perhaps thousands of articles were published claiming proof of something or other because the coefficient was significant at the 5\% level.'' Also: ``The deficiencies in analysis occurred because the focus was on the model and not on the problem.'' Breiman conveys here that the focus was on constructing a model that would faithfully represent the data generation process (if some extremely strong assumptions hold) rather than on creating a model that fits the data well. He implies that the data generation models are linear, writing ``The linear regression model led to many erroneous conclusions that appeared in journal articles waving the 5\% significance level without knowing whether the model fit the data.'' Breiman essentially implies that the hypothesis test calculation of what ``could have'' generated the data (under too strong assumptions) does not reveal what ``definitely did'' generate it, and says almost nothing about the accuracy of the model.

He noted that  statisticians were measuring the performance of these models in-sample (if they measured performance at all), without comparing to other algorithms or performing any type of cross-validation, omissions that sound lethal to any modern statistician or machine learning practitioner. He says: ``For instance, many of the
current application articles in JASA that fit data models have very little discussion of how well their model fits the data.''
In other words, Breiman claims it was a common practice of statisticians to develop models without consideration of either training or test accuracy. A machine learning scientist would never omit these calculations, but a data modeler could. If the data modeler is completely certain they have modeled the data generation process, or at least the variables in it, accurately, then certainly a goodness-of-fit test wouldn't reject the null hypothesis that this model generated the data, or that particular variables were involved in data generation. 

Breiman points out the obvious flaws in this approach: (1) Analysts can never be completely certain they know the form of, or variables involved in, the data generation process, and they can never verify their assumptions in the presence of the Rashomon Effect. Their models may easily be inaccurate or identify the wrong variables and they would not know it.  (2) Variable importance calculated from inaccurate models does not necessarily reflect the importance of variables for the true data generation process. Breiman spends substantial effort explaining the importance of training and cross-validation to data modelers. Breiman's point seems obvious, but his discussion reveals an important divergence of philosophies: the data modelers were so focused on creating guarantees on the data generation process that they missed the possibility that better performance and better understanding could be gained with other models. The algorithmic modelers, conversely, didn't care about data generation or interpretability, and thus chose more complex models, even if simpler ones would have achieved the same performance. This culture clash left no room in between for interpretable algorithmic models.

Thus, while Breiman observed an accuracy/simplicity tradeoff, it could be because the ``sweet spot'' of interpretable algorithmic models was not a goal of either modeling group. If we take only the extremes of overly simplistic data models and overly complex machine learning models, clearly an accuracy/simplicity tradeoff will emerge. For instance, he writes: ``Unfortunately, in prediction, accuracy and simplicity (interpretability) are in conflict. For instance, linear regression gives a fairly interpretable picture of the y, x relation. But its accuracy is usually less than that of the less interpretable neural nets.''


\subsection{CART Trees Are Not Optimized for Accuracy}\label{subsec:cart}

To Breiman's credit, he did attempt to create a middle ground of interpretable algorithmic models. This middle ground is what his 1984 CART algorithm provides \cite{BreimanCART}. CART's models are nonparametric, and decision trees can be quite powerful. However, CART does not actually optimize for any global objective; it instead uses splitting and pruning heuristics. I could not imagine what else could be done on computers from 1984 though.

CART had flaws that Breiman pointed out, the main one being that it does not predict very well. He writes: ``On interpretability, trees rate an A+.'' ... ``While trees rate an A+ on interpretability, they are good, but not great, predictors. Give them, say, a B on prediction.''

Breiman, however, confused CART with the class of decision tree models in this statement -- not every decision tree is a CART tree. He used these terms interchangeably in his quotes above, despite the fact that they are not interchangeable. 
It is possible that a simultaneously interpretable and accurate decision tree exists, but that CART could not find it.
Optimizing to find interpretable yet accurate models requires solving hard computational problems, and Breiman did not have powerful enough computers to investigate this possibility. Nowadays, we can find accurate, sparse, trees within seconds or minutes, even for large datasets \cite{mctavish2022fast,lin2020generalized}. We can even find \textit{all} of the good accurate sparse decision trees \cite{xin2022exploring} for a given dataset, which would have been a practically impossible task on a 1984 computer.

While I grant Breiman some slack for not having the computational power I have access to, I do not think he should have conflated performance on ``decision trees'' with CART performance. He should not have assumed that simply because CART could not find a good decision tree that no such thing exists. He essentially made the same logical flaw that he criticizes data modelers for: assuming that a chosen model (a linear model in the case of data modelers, a CART tree for Breiman) is best in the class, even if there may be a lot of good models out there.

In fact, he made the same mistake twice, stating that support vector machines (SVMs) are more accurate than neural nets: ``These theoretical bounds led to support vector machines (see Vapnik, 1995, 1998) which have proved to be more accurate predictors in classification and regression then neural nets, and are the subject of heated current research.''
We know that this statement is false for non-noisy data, where neural networks have proven their superiority, and SVMs are rarely used. For noisy data, the methods all tend to perform similarly thanks to the Rashomon Effect \cite{SemenovaRuPa2022}.

Breiman invented random forest as a way to handle both problems: they were more accurate, and they were better able to identify causal variables. Here he went completely to the ``dark'' side with his black box algorithm. 

To summarize this section, there is ``simplicity'' as Breiman describes it -- logistic regression models, which were stunted in complexity, or CART models, which were stunted in computation. There is ``simplicity'' from my perspective -- models within a reasonably simple, but still expressive, function class. My models would be fit carefully to data with modern computers, which is how they achieve both performance and interpretability. But I am not the first to claim this. Someone else did, but Breiman did not believe him.

\section{Some People Knew What Breiman Didn't}

One of the commenters on Breiman's article was Bruce Hoadley, working at Fair Isaac and Co. on the design of credit scoring models. Hoadley explained to Breiman almost exactly what I would have said to him. Hoadley observed exactly what the Rashomon Set Theory predicts -- no accuracy/simplicity tradeoff. He observed high-quality performance with interpretable models, including a combination of trees and GAMs, which hit the ``sweet spot'' that Breiman had not found.  Hoadley writes:
``Algorithmic modeling is a very important area of statistics. It has evolved naturally in environments
with lots of data and lots of decisions. But you can do it without suffering the Occam dilemma; for example, use medium trees with interpretable GAMs in the leaves. They are very accurate and
interpretable. And you can do it with data modeling tools as long as you (i) ignore most textbook advice,...'' 


Hoadley also observed the Rashomon Effect: ``What I discovered surprised me. All models fit with anywhere from 27 to 36
characteristics had the same performance on the test sample. This is what Professor Breiman calls `Rashomon and the multiplicity of good models.' ''


Hoadley saw the Rashomon Set Theory in action. He tried many ways to increase the complexity of the models, and saw no increase in performance. In fact, possibly because the GAMs were easier to troubleshoot, the performance of the GAMs actually exceeded that of the black box models.
``To model the interactions, I tried
developing small adjustments on various overlapping segments. No matter how hard I tried, nothing
improved the test sample performance over the global scorecard. I started calling it the Fat Scorecard. Earlier, on this same data set, another Fair, Isaac
researcher had developed a neural network with 2,000 connection weights. The Fat Scorecard slightly outperformed the neural network on the test sample.''

Hoadley was a careful experimenter who observed exactly what theory would predict 20+ years later. Sadly, Breiman simply didn't believe him.
In Breiman's rejoinder to Hoadley, he writes ``His other point of contention is that the Fair, Isaac algorithm retains interpretability, so that it is possible to have both accuracy and interpretability. For clients who like to know what's going on, that's a sellable item. But developments in algorithmic modeling indicate that the Fair, Isaac algorithm is an exception.''
In fact, Breiman was wrong -- it is not the exception, but the norm.

For instance, more recent experiments with newer FICO data have shown a similar lack of tradeoff between accuracy and interpretability \cite{liu2022fast,ChenFICO}.

\section{Causality}
Causality is a much more nuanced topic to discuss than interpretability or simplicity. They are generally separate topics, though interpretability can help understand the data generation process, and  sometimes people do not view models as interpretable unless they have a causal interpretation. Breiman somewhat confuses these topics when he writes ``The goals in statistics are to use data to predict and to get information about the underlying data mechanism.'' This statement is true for some problems, but not for others. The need for causality is problem-dependent. 

It is possible to have a model be interpretable but have no causal interpretation. In predicting criminal recidivism, we use past crimes to predict future ones, but past crimes do not cause someone to commit future crimes, so a causal model does not make sense -- it is only predictive. There are no variables in the dataset that have a causal relationship with the outcome, only correlations. 
There are other problems where causality is central and prediction is secondary. For genetic modeling, we might want to know which gene causes an outcome, and that is the sole goal. Thus, I view causality as being essential to modeling sometimes, but not other times. 

Causality can be ascertained in other ways than fitting a model and examining it. My preferred ways to ascertain causality from observational data are all model-free. 
My lab's latest approach to this problem is the Rashomon Importance Distribution (RID) \cite{donnelly2023the}, which aims to calculate variable importance for the true data generation process.
The RID approach finds many good models (models with low regularized loss), calculated over many bootstraps of the data, to find a probability distribution over each variable's true importance to the data generation process. 

If we want to estimate the effect that a chosen variable $v$ has on the outcome for an individual data point, I prefer Almost Matching Exactly (AME) approaches, which use case-based reasoning \cite{WangEtAlFLAME2021,ParikhRuVo2022,DiengEtAl2019,LannersEtAl23}. In AME, a data point is matched to a group of almost identical points but where the value of $v$ is different. Outcomes from this matched group reveal what would happen to the outcome if $v$ were changed. A recent application of AME methods to pre-trial program evaluation is here \cite{SealeCarlisle2024}. These methods are nonparametric, and there is no data model at all, yet the reasoning processes of these approaches are very interpretable because the matched groups can be inspected. Many algorithmic models may be used to create these estimates, but none of them are used alone to create causal conclusions.

Thus, I veer from Breiman and the commenters entirely to say that we can get causal information about the variables in the data generation process fairly reliably without a data model at all. And, I veer from them (with the exception of Hoadley) on the Occam Dilemma, since it is clear both empirically and theoretically that the accuracy/simplicity tradeoff does not exist in the presence of noisy outcomes. These views place me outside of both of Breiman's Two Cultures.

Clearly, though, I agree with the goals of both cultures: understanding the data generation process and accurate modeling. I just do not think those two goals need to be handled by one single model. Breiman uses interpretability as a way to get to causality, but model interpretability is valuable in its own right, even for models that are not causal. Interpretable notions of causality can be obtained without interpretable models and case-based reasoning methods such as AME can do this.




\section{Conclusion}

The Two Cultures paper has been inspirational for illuminating the important goals of statisticians, including causality, interpretability, and accuracy. I always appreciated Breiman's incisive and perceptive quotes on these goals. Technology and culture have changed a lot since that time, but these goals have not changed. 

While some of Breiman's connections and seeming compromises between these goals seemed obvious to him at the time, the wind has shifted. Specifically, while Breiman discusses data modeling for the purpose of understanding causality, I discussed how causal conclusions can be made more reliably without data models, using the AME techniques. While Breiman views causality as the end goal, it is not necessary for all prediction problems. While Breiman discusses the accuracy/simplicity tradeoff, I discussed how such a tradeoff is not compatible with the Rashomon Effect generally -- assuming ``simple'' models can include generalized additive models and optimized decision trees, not just linear models and CART's unoptimized trees. It is unfortunate that I am not able to discuss these points directly with him. It would also be delightful to show him how powerful modern techniques for interpretable machine learning have become. 

What I would like most to explain to Breiman is how his paper influenced the thinking of generations of scientists. His use of the word ``interpretability'' for inherently interpretable models is proof that Interpretable Machine Learning is much older than what people tend to think; it is certainly older than the more recent area of ``Explainability'' (i.e., explaining black boxes). If one includes CART, interpretable machine learning dates back to at least 1984.


I mentioned earlier that I thought Breiman knew that interpretable models were valuable outside of causal interpretations. The following quote from Breiman reveals it, because models doctors use do not try to model the data generation process; they try to make good predictions on each patient. 
Breiman writes:
``My biostatistician friends tell me, `Doctors can interpret logistic regression.' There is no way they can interpret a black box containing fifty trees hooked together. In a choice between accuracy and interpretability, they’ll go for interpretability.'' I completely agree. Thank you, Leo Breiman!

\section*{Acknowledgments}
Thank you to my incredible co-authors who have helped shape my view on these topics across the years. Particular thanks to Ron Parr, Panyu Chen, and Chudi Zhong for comments.




\pagestyle{empty}
\renewcommand{\bibname}{\large{\textbf{Bibliography}}}
\bibliographystyle{alpha}
\bibliography{ref}

\end{document}